\documentclass{article} % For LaTeX2e
\usepackage[final]{colm2025_xllmworkshop}

\usepackage{microtype}
\usepackage{hyperref}
\usepackage{url}
\usepackage{booktabs}
\usepackage{enumitem}
\usepackage{graphicx}
\usepackage{float}
\usepackage{lineno}
\usepackage{subcaption}

\definecolor{darkblue}{rgb}{0, 0, 0.5}
\hypersetup{colorlinks=true, citecolor=darkblue, linkcolor=darkblue, urlcolor=darkblue}

\title{Reasoning Riddles: How Explainability Reveals Cognitive Limits in Vision-Language Models}

\author{Prahitha Movva \\
University of Massachusetts Amherst \\
Amherst, MA 01003, USA \\
\texttt{prahitha.movva03@gmail.com} \\
}

\begin{document}

\ifcolmsubmission
\linenumbers
\fi

\maketitle

\begin{abstract}
Vision-Language Models (VLMs) excel at many multimodal tasks, yet their cognitive processes remain opaque on complex lateral thinking challenges like rebus puzzles. While recent work has demonstrated these models struggle significantly with rebus puzzle solving, the underlying reasoning processes and failure patterns remain largely unexplored. We address this gap through a comprehensive explainability analysis that moves beyond performance metrics to understand how VLMs approach these complex lateral thinking challenges. Our study contributes a systematically annotated dataset of 221 rebus puzzles across six cognitive categories, paired with an evaluation framework that separates reasoning quality from answer correctness. We investigate three prompting strategies designed to elicit different types of explanatory processes and reveal critical insights into VLM cognitive processes. Our findings demonstrate that reasoning quality varies dramatically across puzzle categories, with models showing systematic strengths in visual composition while exhibiting fundamental limitations in absence interpretation and cultural symbolism. We also discover that prompting strategy substantially influences both cognitive approach and problem-solving effectiveness, establishing explainability as an integral component of model performance rather than a post-hoc consideration.
\end{abstract}

\section{Introduction}

The ability to solve rebus puzzles—visual-textual riddles that encode phrases through symbolic representations—requires a sophisticated integration of visual perception, symbolic interpretation, and linguistic creativity. These puzzles challenge both human and artificial intelligence by demanding that solvers recognize visual patterns, understand symbolic relationships, and bridge between literal and metaphorical meanings.

Recent work has established that VLMs face significant challenges when solving rebus puzzles, with even state-of-the-art models achieving limited success on these visual wordplay tasks \citep{lee2025puzzledpuzzlesvisionlanguagemodels, gritsevskiy2024rebusrobustevaluationbenchmark}. While these performance-focused studies have revealed the extent of VLM limitations on lateral thinking challenges, a critical gap remains: we lack understanding of how these models approach these tasks and why they fail.

This opacity in reasoning processes becomes particularly problematic as these systems are increasingly deployed in applications requiring transparent decision-making. While prior work measured VLM performance on rebus puzzles, we investigate how models reason and why they fail through systematic explainability analysis. This shift from performance evaluation to process analysis represents a crucial step toward understanding the cognitive mechanisms underlying VLM behavior on complex multimodal inference tasks.

Rebus puzzles represent an ideal testbed for investigating explainability in complex reasoning scenarios. Solving a rebus puzzle requires the integration of multiple cognitive skills to synthesize the components into coherent solutions. This multi-faceted nature of rebus puzzles makes them ideal for explainability research, as they require models to articulate not just what they see, but how they transform visual and textual cues into abstract meanings.

% Unlike straightforward visual recognition tasks where failure modes are often apparent, complex reasoning scenarios involving spatial relationships, abstract thinking, and cultural knowledge require deeper investigation to understand when and why models struggle. The ability to interpret and explain VLM reasoning processes is not merely academically interesting—it is essential for developing more robust, reliable, and trustworthy AI systems.

Our work makes several key contributions to multimodal explainability research:

\begin{itemize}[itemsep=0.2pt, topsep=0pt]
    \item demonstrate that prompting strategy fundamentally affects both reasoning and problem-solving effectiveness, establishing explainability as an integral component of model performance rather than an external consideration.
    \item contribute a systematically annotated dataset designed specifically for explainability research, with puzzles categorized across cognitive dimensions.
    \item provide actionable insights by identifying specific reasoning failure patterns.
\end{itemize}

% Our open-source dataset and evaluation methodology will enable continued research into VLM explainability, supporting the broader goal of developing more transparent and trustworthy AI systems.

\section{Related work}

Multimodal reasoning benchmarks have evolved to assess increasingly complex cognitive abilities. PuzzleWorld \citep{li2025puzzleworldbenchmarkmultimodalopenended} established a comprehensive framework with 667 puzzlehunt-style problems designed to assess step-by-step, open-ended, and creative multimodal inference. This work emphasized the importance of detailed reasoning traces and cognitive skill labels for understanding model capabilities. Building on this foundation, the REBUS benchmark \citep{gritsevskiy2024rebusrobustevaluationbenchmark} focused specifically on visual-textual wordplay, providing 333 original examples that challenge models to decode symbolic representations across diverse categories. The benchmark revealed that even advanced models like GPT-4V struggle with the symbolic interpretation required for rebus puzzles, achieving only modest performance levels.

The assessment of lateral thinking in AI has emerged to be critical, as standard benchmarks may not capture the full spectrum of human cognitive abilities. Lateral thinking—characterized by indirect, creative problem-solving—poses challenges for current evaluation methodologies. RiddleSense \citep{lin2021riddlesensereasoningriddlequestions} demonstrated that models struggle with inference beyond explicitly stated information. BRAINTEASER \citep{jiang2023brainteaserlateralthinkingpuzzles} focuses on puzzles requiring departure from conventional logic, while LatEval \citep{huang2024latevalinteractivellmsevaluation} emphasizes interactive inquiry and creative problem-solving.

Beyond formal benchmarks, recent explorations have begun to assess model creativity through abstract visual tasks that mirror lateral thinking challenges. For instance, \citet{Jun2024airebus} proposes using tasks like rebus puzzles and pattern completions to measure models' capacity for symbolic abstraction and metaphorical thinking, emphasizing the link between explainability and emergent creativity in multimodal inference. These benchmarks have revealed that models often default to conventional solution patterns even when creative approaches are required. \citet{lee2025puzzledpuzzlesvisionlanguagemodels}'s work on rebus puzzles revealed that while models show competency in direct visual-text alignment tasks, they achieve limited success when puzzles demand symbolic abstraction, phonetic manipulation, and cultural context understanding. 

Recent advances in multimodal explainability have explored various approaches to understanding model reasoning. Textual explanations, including Chain-of-Thought (CoT) prompting \citep{wei2023chainofthoughtpromptingelicitsreasoning, zhang2024improvevisionlanguagemodel}, have shown promise in making model reasoning more transparent, though their reliability remains questionable \cite{chen2024measuringimprovingchainofthoughtreasoning}. However, most explainability research has focused on relatively straightforward tasks, leaving a significant gap in our understanding of how these architectures approach complex lateral thinking challenges.
% The recent demonstration of VLM limitations on rebus puzzles (Chan et al., 2025) highlights this explainability gap. While we now understand that models struggle with these tasks, we lack systematic frameworks for analyzing why they struggle and how their reasoning processes unfold. This gap is particularly problematic because complex reasoning failures often stem from subtle cognitive bottlenecks that cannot be identified through performance metrics alone.

% A key limitation in current VLM explainability research is the lack of systematic evaluation frameworks for assessing explanation quality in complex reasoning scenarios. While several metrics have been proposed for evaluating textual explanations in language models (Narang et al., 2020), these often fail to capture the unique challenges of multimodal reasoning where visual and linguistic elements must be coherently integrated, and where reasoning quality may diverge significantly from answer correctness.

% Building on recent performance evaluations that have established the difficulty of rebus puzzles for VLMs, we shift focus from what models can do to how they reason and why they fail. By emphasizing reasoning process analysis rather than mere performance metrics and developing evaluation frameworks that separate reasoning quality from answer correctness, we provide essential insights for understanding the cognitive mechanisms underlying VLM behavior on complex multimodal reasoning challenges.

\section{Dataset collection and annotation}

We constructed a dataset of 221 rebus puzzles from three sources (Rainiers Family Instagram account \footnote{\url{https://www.instagram.com/rainiersfamily/}}, Reader's Digest \footnote{\url{https://www.rd.com/list/rebus-puzzles/}}, and Rebus Puzzles subreddit \footnote{\url{https://www.reddit.com/r/rebus/}}), ensuring broad coverage of puzzle types, difficulty levels, and cultural contexts. For more details about the dataset distributions, see Appendix \ref{app:dataset-dist}.

We annotated each puzzle with its category and theme. The cognitive categorization scheme includes six distinct categories, namely, Spatial Encoding (SE), Absence Reasoning (AR), Quantitative Logic (QL), Cultural Symbolism (CS), Phonetic Transformation (PT), and, Visual Composition (VC). A detailed description of each of the categories is given in the Appendix \ref{app:puzzle-cat}. The thematic annotation captures the content domain of puzzle solutions across six categories: food and cuisine, movies and entertainment, music and songs, proverbs and sayings, idioms and expressions, and common phrases. This thematic categorization enables analysis of whether VLM performance varies based on the cultural and conceptual domains being tested.

% Spatial encoding puzzles require understanding text positioning, orientation, or spatial relationships within the visual layout. Absence reasoning puzzles involve missing elements, negation symbols, or crossed-out words that must be interpreted as conceptual absence. Quantitative logic puzzles incorporate mathematical operations, counting, or numerical relationships. Cultural symbolism puzzles rely on metaphors, idioms, pop culture references, or culturally-specific knowledge. Phonetic transformation puzzles require sound-based wordplay, homophones, or pronunciation-based reasoning. Visual composition puzzles combine multiple visual elements that must be integrated to form the solution.

% Our final dataset comprises 222 rebus puzzles with the following distribution across cognitive categories: spatial encoding (45 puzzles, 20.3\%), absence reasoning (38 puzzles, 17.1\%), quantitative logic (32 puzzles, 14.4\%), cultural symbolism (41 puzzles, 18.5\%), phonetic transformation (35 puzzles, 15.8\%), and visual composition (31 puzzles, 13.9\%).
To ensure annotation quality and consistency, we implemented a rigorous quality control process. All puzzles were initially annotated by the primary researcher, followed by a validation phase where a subset of 50 puzzles (22.5\% of the dataset) was independently reviewed by two additional annotators familiar with puzzle-solving and cognitive categorization. Inter-annotator agreement was measured using Cohen's kappa, achieving $\kappa \ge 0.91$ for both cognitive and thematic categories, indicating strong agreement and annotation reliability.

\section{Methodology}

\subsection{Prompting strategies}

We designed three distinct prompting strategies—explain-then-solve (ETS), solve-then-explain (STE), and component-guided (CG)—each targeting specific aspects of the reasoning process and explanation generation. ETS asks models to first describe visual elements, then explain relationships, before solving. STE reverses this order, requiring the answer first followed by justification. CG provides explicit category and theme labels to scaffold problem-solving. Detailed prompt descriptions appear in Appendix \ref{app:prompts}.

% The explain-then-solve strategy encourages models to engage in explicit step-by-step reasoning before providing their final answer. The prompt instructs models to first analyze the visual and textual elements present in the puzzle, identify potential wordplay or symbolic relationships, consider spatial arrangements and visual cues, and then synthesize these observations into a coherent solution. This approach aims to capture the model's reasoning process as it unfolds, potentially revealing how different cognitive skills are integrated during problem-solving.
% The solve-then-explain strategy promotes post-hoc rationalization by first asking models to provide their answer and then explain their reasoning. The prompt requests the model to state their solution immediately, then describe the reasoning steps that led to that conclusion. This approach tests whether models can provide coherent explanations for their solutions and helps distinguish between genuine reasoning and pattern matching.
% The component-guided prompting strategy provides categorical and thematic scaffolding to support model reasoning. In addition to the puzzle image, models receive explicit information about the puzzle's cognitive category (e.g., "This puzzle involves spatial encoding") and thematic domain (e.g., "The answer relates to food"). This approach tests whether additional context can improve both reasoning quality and explanation coherence, while also revealing how models utilize explicit guidance in their problem-solving process.

\subsection{Evaluation framework}
We evaluated our methodology using state-of-the-art vision-language systems. GPT-o3 represents the current state-of-the-art among commercial offerings, while Claude Opus-4 and Sonnet-4 provide important comparison points across different capability levels within the same model family. We initially explored open-source alternatives such as InternVL and Qwen2.5 VL; however, preliminary testing revealed significant performance degradation even on straightforward examples, suggesting these models may not yet possess the baseline capabilities required for meaningful analysis of complex lateral thinking tasks. Given our focus on understanding cognitive processes in capable systems, we prioritized models that could successfully solve a substantial portion of puzzles, enabling richer analysis of both successes and failures.

We ensure consistent evaluation conditions across all prompting strategies and puzzle categories, with each model receiving identical puzzle presentations and prompt variations. We conducted all evaluations using identical computational environments and model configurations. We processed puzzles in randomized order to minimize potential ordering effects and conducted multiple runs per puzzle to assess consistency. The manual evaluation of solution quality was performed by trained evaluators following detailed rubrics developed specifically for rebus puzzle assessment. 

Our evaluation framework moves beyond simple answer accuracy to provide a fine-grained assessment of solution quality along four key dimensions: correctness, coherence, completeness, and cognitive skill use. For each model response, trained evaluators independently rated these dimensions on a standardized 5-point scale.

\begin{itemize}[itemsep=0.2pt, topsep=0pt]
\item Correctness captures whether the final answer accurately solves the puzzle.
\item Coherence evaluates the logical consistency and flow of the reasoning process.
\item Completeness measures the extent to which the explanation accounts for all relevant elements of the puzzle.
\item Cognitive Skill Use assesses whether the model applies the appropriate cognitive approach—such as spatial understanding, phonetic manipulation, or cultural inference—based on the puzzle category.
\end{itemize}

This evaluation framework enables a more nuanced understanding of model behavior, highlighting how prompting strategies influence not just final answers but the underlying solution pathways. It also facilitates the identification of systematic strengths and failure modes across different cognitive challenge types.

\section{Results}

Our evaluation reveals significant variations in VLM performance across different prompting strategies, with important implications for interpretability research. Table \ref{tab:correctness} summarizes the correctness rates for each model across the three prompting strategies. GPT-o3 consistently outperforms the other models, with a slight edge for the component-guided strategy. Notably, all three models show improved correctness with the component-guided approach, supporting the hypothesis that structured cognitive scaffolding can enhance problem-solving performance. These trends highlight the impact of prompting design on model effectiveness in lateral thinking tasks.

\begin{table}[h]
\centering
\begin{tabular}{lccc}
\toprule
\textbf{Model} & \textbf{ETS} & \textbf{STE} & \textbf{CG} \\
\midrule
\textbf{GPT-o3}     & 76.5\% & 76.0\% & 77.4\% \\
\textbf{Claude Opus-4}     & 50.7\% & 42.1\%  & 62.0\% \\
\textbf{Claude Sonnet-4}   & 40.7\% & 30.8\%  & 45.7\% \\
\bottomrule
\end{tabular}
\caption{Correctness percentages for each model across prompting strategies.}
\label{tab:correctness}
\end{table}

Figure 4 (Appendix \ref{app:eval-metrics}) shows reasoning quality scores across all evaluation dimensions.

\subsection{Category-wise analysis}

Analysis across cognitive categories reveals significant variations in model capabilities. Visual composition (77\% average correctness) and spatial encoding (73\%) yielded the strongest performance, while absence reasoning (23\%) emerged as the most challenging category. Models consistently struggled to interpret crossed-out text, missing elements, and negation symbols as meaningful absence concepts. This limitation appears fundamental rather than superficial, suggesting gaps in abstract reasoning capabilities when dealing with implicit or negative information.

\subsection{Cognitive complexity analysis}

Our most critical finding emerges from analyzing performance across cognitive complexity levels:

\begin{table}[h]
\centering
\begin{tabular}{lccc}
\toprule
\textbf{Category Count} & \textbf{Puzzle Count} & \textbf{Average Correctness} & \textbf{Performance Drop} \\
\midrule
\textbf{1 category}     & 123 & 70.2\% & - \\
\textbf{2 categories}   & 80 & 53.4\%  & -15.8\% \\
\textbf{3 categories}   & 17  & 41.9\%  & -26.3\% \\
\textbf{4 categories}   & 1  & 25.0\%  & -43.2\% \\
\bottomrule
\end{tabular}
\caption{Average correctness for CG prompting across all models.}
\label{tab:complexity}
\end{table}

Table \ref{tab:complexity} shows that accuracy degrades systematically as cognitive complexity increases, suggesting fundamental limitations in parallel cognitive processing rather than simple difficulty scaling. This degradation suggests that models often focus narrowly on a single category while failing to integrate multiple cognitive strategies. When multiple strategies are simultaneously required, models show confusion or fixation, indicating limitations in coordinating parallel skills.

\subsection{Analysis of errors}

Common failure patterns include fixation on surface visual elements without considering deeper symbolic interpretations, failure to consider multiple possible interpretations of ambiguous elements, and inadequate integration of cultural or contextual knowledge required for solution. As illustrated in Appendix \ref{app:error-analysis} (Figures \ref{fig:cultural-symbolism}, \ref{fig:absence-reasoning}, and \ref{fig:skill-fixation}), these failure patterns often reflect single-skill fixation, misperception of absent elements, or cultural inference gaps. This highlights that our error taxonomy is not just anecdotal but systematic across multiple puzzle types.

\section{Findings}

Our findings extend beyond performance metrics to reveal critical cognitive bottlenecks. The superior performance on visual composition tasks suggests that these systems have developed robust mechanisms for integrating multiple visual elements, likely reflecting effective cross-modal attention mechanisms. However, the poor performance on absence reasoning indicates not just performance limitations but fundamental gaps in abstract conceptual processing—these models struggle not only to solve these puzzles but to articulate coherent explanations about implicit information and negation. These results align with the qualitative error patterns in Appendix \ref{app:error-analysis}, reinforcing that absence reasoning and cultural symbolism remain consistent cognitive bottlenecks. The superior performance of component-guided prompting demonstrates that encouraging explicit reasoning processes can enhance problem-solving capabilities, supporting theories that transparency and effectiveness are inherently linked rather than separate concerns.

\section{Limitations}

Our study identifies several limitations in current multimodal capabilities and interpretability. The ability to reason about implicit information, negation, and conceptual absence appears crucial for robust cognitive systems. Cultural knowledge gaps present another significant challenge, with model performance varying dramatically based on the cultural specificity of puzzle content.

Another practical limitation concerns computational cost-benefit tradeoffs across prompting strategies. CG prompting generates substantially longer explanations (approximately 2-3x ETS responses) but yields modest accuracy gains: less than 1\% for GPT-o3, approximately 5\% for Sonnet-4, and 11\% for Opus-4. This suggests explicit cognitive scaffolding benefits lower-capability models more substantially, while high-performing models may achieve better efficiency with simpler strategies.

The inconsistency between reasoning quality and answer accuracy across different prompting strategies indicates that current evaluation approaches may be insufficient for assessing true cognitive capabilities. The development of more sophisticated evaluation frameworks that can distinguish between genuine reasoning and pattern matching represents an important methodological challenge for the field.

\section{Future Work}

\subsection{Dataset}

Our study highlights multiple avenues to improve both the evaluation of multimodal reasoning and the design of more transparent models. On the dataset side, we plan to expand beyond 500 puzzles and diversify sources to capture richer cultural, linguistic, and visual phenomena. We also envision extending to lateral thinking tasks such as wordplay riddles, visual logic puzzles, and quantitative teasers, thereby creating a more comprehensive suite for reasoning evaluation. Interactive puzzle-solving settings, where intermediate hypotheses and backtracking are logged, could further support fine-grained analysis.

On the analysis side, we will explore concept-based interpretability methods to understand what semantic features models rely on. Recent work has shown that VLM embedding spaces can be decomposed into sparse, human-interpretable concept vectors using sparse autoencoders or concept embeddings \citep{NEURIPS2024_996bef37}. Applying such methods to rebus puzzles could reveal whether models activate on the expected concepts (e.g., negation mark, homophone, idiom) or whether their reasoning reflects spurious shortcuts. Similarly, architectures like STAIR \citep{chen-etal-2023-stair} demonstrate that aligning images and texts into a shared sparse token space can enhance interpretability without hurting performance, offering a potential template for reasoning-specific architectures.

On the modeling side, integrating structured reasoning modules could strengthen puzzle solving. Programmatic approaches such as ViperGPT \citep{surís2023vipergptvisualinferencepython} and GENOME \citep{chen2023genomegenerativeneurosymbolicvisual} show that decomposing problems into executable steps or modular skills yields both higher accuracy and interpretable reasoning traces. For our setting, a neuro-symbolic or modular extension could allow explicit handling of categories like absence interpretation or phonetic transformation, which remain core weaknesses of current systems. Together, these directions aim not only to improve performance but also to bridge evaluation, interpretability, and architectural design for multimodal intelligence.

\section{Conclusion}

We introduced a new benchmark of 221 rebus puzzles spanning six cognitive categories, designed to probe reasoning alignment in vision-language models. Our experiments across prompting strategies and models reveal clear cognitive bottlenecks: while visual composition is handled relatively well, absence reasoning and culturally grounded puzzles remain particularly challenging. We further identified common failure modes—such as surface fixation, phonetic drift, and neglect of negation—that highlight both data and modeling gaps.

Beyond establishing baseline results, our work underscores the dual need for better interpretability and stronger reasoning architectures. Transparent evaluations, enriched with human comparisons and concept-level probes, will help clarify whether models genuinely understand puzzle components or merely exploit superficial cues. Likewise, modular or programmatic reasoning approaches offer promising avenues to scaffold the multi-step logic that riddles demand. We view rebus puzzles as a fertile testbed where explainability, dataset design, and model innovation intersect, and we hope this benchmark catalyzes progress toward models that reason more like humans—not only in performance, but in how their reasoning can be inspected and trusted.

% This work advances understanding of VLM reasoning and explainability through lateral thinking challenges. Using rebus puzzles as a multimodal testbed, we demonstrated the value of process-oriented evaluation over simple performance metrics and identified specific cognitive strengths and limitations in current VLM architectures.
% Our findings reveal that explainability enhances rather than merely illuminates problem-solving capabilities. The superior performance of explain-then-solve prompting and improved reasoning quality with component-guided approaches show that transparency mechanisms are integral to effective reasoning, not just desirable features.

% The open-source release of our dataset enables continued VLM explainability research, supporting development of more transparent, reliable, and trustworthy AI systems. As VLMs deploy in critical applications, understanding and improving their reasoning processes becomes practically essential. 
% Our work provides actionable insights for VLM development, evaluation, and deployment. 
% Identified reasoning bottlenecks offer targeted improvement directions, while demonstrated prompting strategy importance provides immediate practical guidance for maximizing performance in reasoning-intensive applications. This comprehensive analysis contributes to broader AI capability understanding while establishing methodological foundations for continued explainable AI research.

\section*{Acknowledgments}
We thank the creators of the Rainier Family Instagram account, Reader's Digest puzzle collections, and the r/rebuspuzzles community for providing the puzzle content that made this research possible. We also acknowledge the volunteer annotators who contributed to our dataset validation process. The dataset introduced in this paper will be made available at: \href{https://github.com/Prahitha/rebus-puzzles}{https://github.com/Prahitha/rebus-puzzles}

\bibliography{colm2025_conference}
\bibliographystyle{colm2025_conference}

\appendix
\section{Appendix}

\subsection{Dataset distribution}
\label{app:dataset-dist}

\begin{table}[H]
\centering
\begin{tabular}{lc}
\toprule
\textbf{Source} & \textbf{Number of Puzzles} \\
\midrule
Rainier's Instagram & 181 \\
Reader's Digest      & 28 \\
Reddit (r/rebus)     & 12 \\
\midrule
\textbf{Total}       & \textbf{221} \\
\bottomrule
\end{tabular}
\caption{Distribution of rebus puzzles by source. The dataset includes a total of 221 puzzles curated from diverse online repositories.}
\label{tab:dataset-sources}
\end{table}

\begin{table}[h]
\centering
\begin{tabular}{p{10.2cm}cc}
\toprule
\textbf{Category Combination} & \textbf{Count} & \textbf{\%} \\
\midrule
Spatial Encoding & 62 & 28.1 \\
Visual Composition & 43 & 19.5 \\
Spatial Encoding, Visual Composition & 16 & 7.2 \\
Cultural Symbolism, Visual Composition & 15 & 6.8 \\
Quantitative Logic & 13 & 5.9 \\
Phonetic Transformation, Spatial Encoding & 10 & 4.5 \\
Phonetic Transformation, Quantitative Logic & 10 & 4.5 \\
Quantitative Logic, Visual Composition & 6 & 2.7 \\
Cultural Symbolism, Spatial Encoding, Visual Composition & 6 & 2.7 \\
Quantitative Logic, Spatial Encoding & 5 & 2.3 \\
Cultural Symbolism, Spatial Encoding & 5 & 2.3 \\
Absence Reasoning, Visual Composition & 4 & 1.8 \\
Phonetic Transformation, Quantitative Logic, Visual Composition & 3 & 1.4 \\
Cultural Symbolism, Phonetic Transformation, Spatial Encoding & 3 & 1.4 \\
Absence Reasoning & 3 & 1.4 \\
Cultural Symbolism, Quantitative Logic & 3 & 1.4 \\
Absence Reasoning, Spatial Encoding & 2 & 0.9 \\
Phonetic Transformation, Visual Composition & 2 & 0.9 \\
Absence Reasoning, Phonetic Transformation, Visual Composition & 2 & 0.9 \\
Phonetic Transformation & 2 & 0.9 \\
Cultural Symbolism, Quantitative Logic, Visual Composition & 1 & 0.5 \\
Cultural Symbolism, Phonetic Transformation, Visual Composition & 1 & 0.5 \\
Absence Reasoning, Phonetic Transformation & 1 & 0.5 \\
Absence Reasoning, Quantitative Logic & 1 & 0.5 \\
Phonetic Transformation, Quantitative Logic, Spatial Encoding & 1 & 0.5 \\
Cultural Symbolism, Quantitative Logic, Spatial Encoding, Visual Composition & 1 & 0.5 \\
\midrule
\textbf{Total} & 221 & 100.0 \\
\bottomrule
\end{tabular}
\caption{Distribution of Category Combinations in Rebus Puzzles}
\label{tab:category_combinations_distribution}
\end{table}

\subsection{Puzzle categories}
\label{app:puzzle-cat}

\begin{table}[H]
\centering
\begin{tabular}{p{4.5cm}p{8.5cm}}
\toprule
\textbf{Category} & \textbf{Description} \\
\midrule
Spatial Encoding (SE) & Text positioning or orientation matters; e.g., The word \textit{"GET"} written above the word \textit{"IT"} = \textit{"get over it"} \\
Absence Reasoning (AR) & Puzzles involving missing elements, negation, or crossed-out words \\
Quantitative Logic (QL) & Involves mathematical or counting operations; e.g., \textit{"50\% and 5/10 = Half and Half"}-type puzzles \\
Cultural Symbolism (CS) & Relies on metaphors, idioms, or cultural references including pop culture or region-specific expressions \\
Phonetic Transformation (PT) & Sound-based wordplay or homophones \\
Visual Composition (VC) & Puzzles requiring the integration of multiple visual elements to convey a phrase or concept \\
\bottomrule
\end{tabular}
\caption{Descriptions of cognitive categories used for rebus puzzle annotation.}
\label{tab:cognitive-categories}
\end{table}

\subsection{Prompts}
\label{app:prompts}

\begin{table}[H]
\centering
\begin{tabular}{p{3.5cm}p{9.5cm}}
\toprule
\textbf{Prompting Strategy} & \textbf{Prompt Template (with structure)} \\
\midrule
\textbf{Explain-then-Solve} & 
Look at this rebus puzzle image carefully. First, describe exactly what you see (text, images, positioning, colors, etc.). Then, explain how these visual elements relate to each other. Finally, provide your solution to the puzzle. \\
& \textbf{Format:}

VISUAL DESCRIPTION: [what you see]

REASONING: [how elements connect]

SOLUTION: [final answer] \\
\midrule
\textbf{Solve-then-Explain} & 
Solve this rebus puzzle and provide the answer, then explain your reasoning process. \\
& \textbf{Format:}

SOLUTION: [final answer]

EXPLANATION: [detailed reasoning for why this is correct] \\
\midrule
\textbf{Component-Guided} & 
Consider the category: [category]

Consider the theme: [theme]

Analyze this rebus puzzle by addressing each component:

1. Visual elements (text, images, symbols)

2. Spatial relationships (positioning, orientation)

3. Cultural/linguistic context needed

4. Solution derivation

FINAL ANSWER: [solution] \\
\bottomrule
\end{tabular}
\caption{Prompting strategies and templates used across all experiments.}
\label{tab:prompt-templates}
\end{table}

\subsection{Error analysis}
\label{app:error-analysis}

\subsubsection{Cultural symbolism failures}

\begin{figure}[H]
\centering
\begin{subfigure}{0.45\linewidth}
    \includegraphics[width=\linewidth]{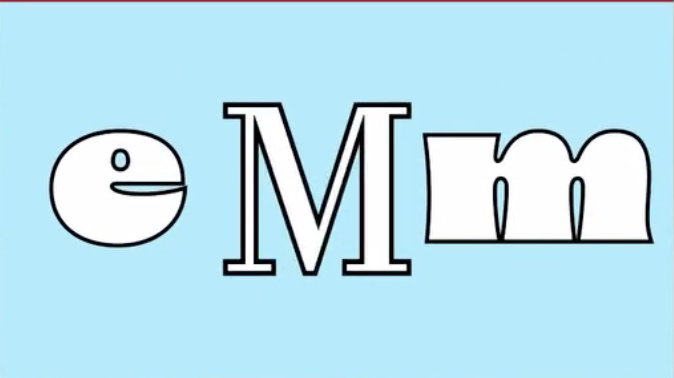}
    \caption{'Eminem' puzzle requiring recognition of multiple font styles and phonetic mapping.}
    \label{fig:eminem}
\end{subfigure}
\hfill
\begin{subfigure}{0.45\linewidth}
    \includegraphics[width=\linewidth]{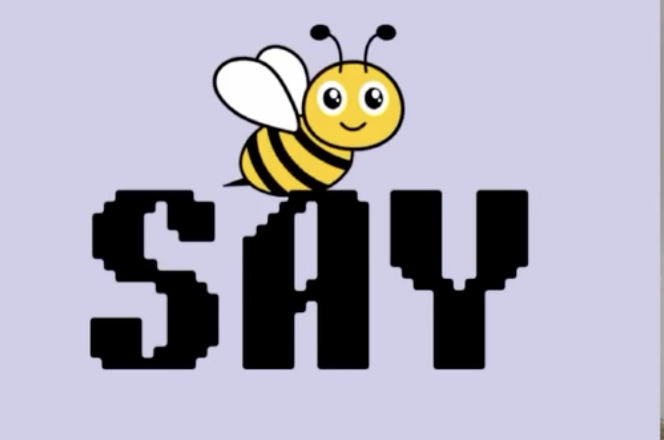}
    \caption{Beyoncé: A bee over the word say}
    \label{fig:beyonce}
\end{subfigure}

\caption{Examples for errors in cultural symbolism.}
\label{fig:cultural-symbolism}
\end{figure}

Cultural symbolism puzzles require models to bridge visual and phonetic reasoning with domain-specific knowledge, revealing systematic limitations in how VLMs access and apply cultural context. Figure \ref{fig:eminem} exemplifies this challenge through its multi-layered encoding: solvers must recognize two distinct letter forms (em, M), font variations, and phonetically map "e-M-m" to the rapper's stage name. GPT-o3 successfully solves this across all prompting strategies. However, Claude Sonnet exhibits complete failure, instead interpreting the visual as "MEDIUM" (fixating on letter size relationships) or "Time" (hallucinating a clock face in the circular 'e'). Claude Opus shows partial success, solving correctly only under the CG condition, which indicates that explicit cognitive scaffolding can activate the correct reasoning pathway but the model lacks autonomous strategy selection.

Figure \ref{fig:beyonce} requires recognizing a cartoon bee, identifying the text "SAY," understanding the spatial relationship ("on"), and phonetically mapping "bee-on-say" to the celebrity name "Beyoncé." Both Claude models consistently misidentify "SAY" as "SHY," demonstrating fundamental perceptual errors that cascade through subsequent reasoning. More critically, even when explicitly corrected that the text reads "SAY," Claude Opus recognizes the phonetic pattern "bee-on-say" but answers "essay" instead of "Beyoncé." The model correctly performs phonetic matching but cannot or will not complete the final cultural inference. GPT-o3 solves this puzzle successfully across all strategies, demonstrating that the required capabilities exist in current VLMs but are not uniformly accessible.
% These patterns indicate that cultural symbolism failures stem not from missing knowledge but from systematic biases in how models prioritize and integrate different reasoning types.

\subsubsection{Absence reasoning failures}

\begin{figure}[H]
\centering
\begin{subfigure}{0.45\linewidth}
    \includegraphics[width=\linewidth]{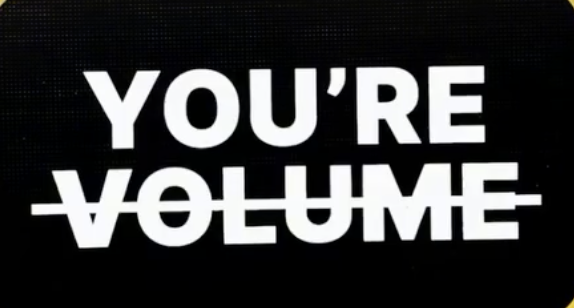}
    \caption{'You're on mute' requires interpreting crossed-out text as negation.}
    \label{fig:mute}
\end{subfigure}
\hfill
\begin{subfigure}{0.45\linewidth}
    \includegraphics[width=\linewidth]{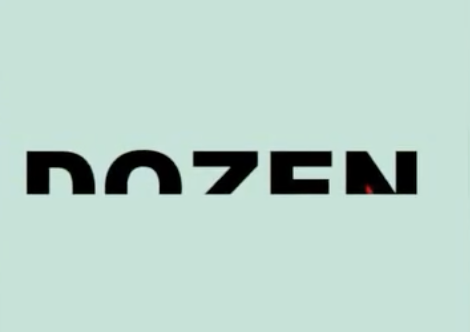}
    \caption{Half a dozen: Top half of the word DOZEN}
    \label{fig:dozen}
\end{subfigure}

\caption{Examples for errors in absence reasoning.}
\label{fig:absence-reasoning}
\end{figure}

Absence reasoning represents the most severe cognitive bottleneck in our evaluation, with models achieving only 23\% average correctness on puzzles requiring interpretation of missing elements, negation, or crossed-out text. Figure \ref{fig:mute} requires recognizing that crossed-out "VOLUME" signifies absence of sound, mapping this to the concept "mute," and constructing the culturally relevant phrase "You're on mute." GPT-o3 solves this correctly across all strategies. However, both Claude models exhibit systematic misinterpretation patterns. Sonnet alternates between "You're quiet" (CG), "You're loud" (STE), and fails to construct any coherent phrase (ETS). Opus shows similar instability, proposing "You're out of volume," "You're welcome", and "You're quiet."

Figure \ref{fig:dozen} shows only the top half of the word "DOZEN," requiring solvers to recognize partial text, infer the complete word, and understand that showing half of "DOZEN" represents the phrase "half a dozen." Remarkably, all models completely misperceive the visual. GPT-o3 hallucinates nonexistent characters across all strategies: "NO7EN" as "FROZEN" (ETS), "RO7EN" with embedded "F" as "FROZEN" (STE), and "SE7EN" as "SEVEN" (CG). Claude models perceive complete "DOZEN" without recognizing truncation, proposing "ELEVEN" (dozen minus one), and "DIRTY DOZEN" (bold letters as "dirty"). This suggests that the failure mode is perceptual rather than purely reasoning-based—models appear to "see" complete alternative texts rather than recognizing visual incompleteness. Whether this stems from architectural limitations in visual encoding, overly aggressive pattern completion in early processing stages, or insufficient training on partially occluded text remains an open question requiring further investigation. However, the practical implication is clear that current VLMs demonstrate systematic unreliability when tasks require precise attention to what is and isn't visually present.

\subsubsection{Single skill fixation}

\begin{figure}[H]
\centering
\begin{subfigure}{0.45\linewidth}
    \includegraphics[width=\linewidth]{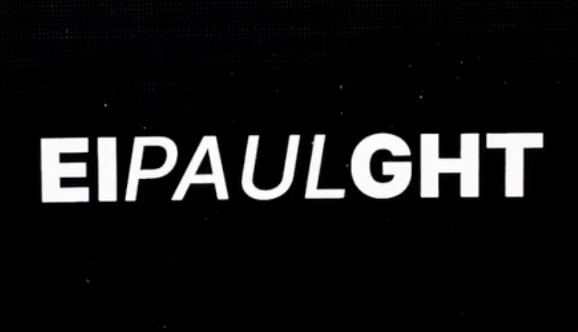}
    \caption{Pollinate: The word PAUL in a different font and in the middle of the word EIGHT. Paul in Eight which sounds like Pollinate}
    \label{fig:pollinate}
\end{subfigure}
\hfill
\begin{subfigure}{0.45\linewidth}
    \includegraphics[width=\linewidth]{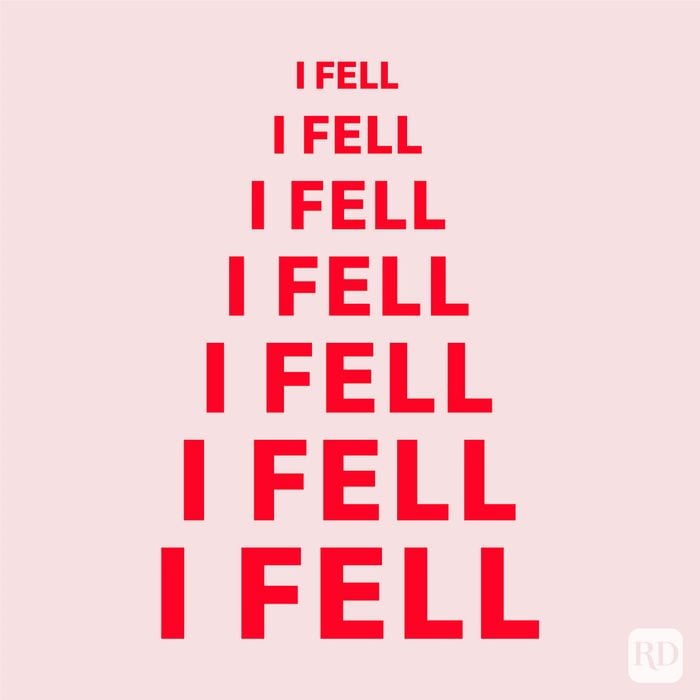}
    \caption{Eiffel Tower: The words I FELL are arranged in the shape of a tower on top of each other}
    \label{fig:eiffel-tower}
\end{subfigure}

\vspace{0.5cm}

\begin{subfigure}{0.45\linewidth}
    \includegraphics[width=\linewidth]{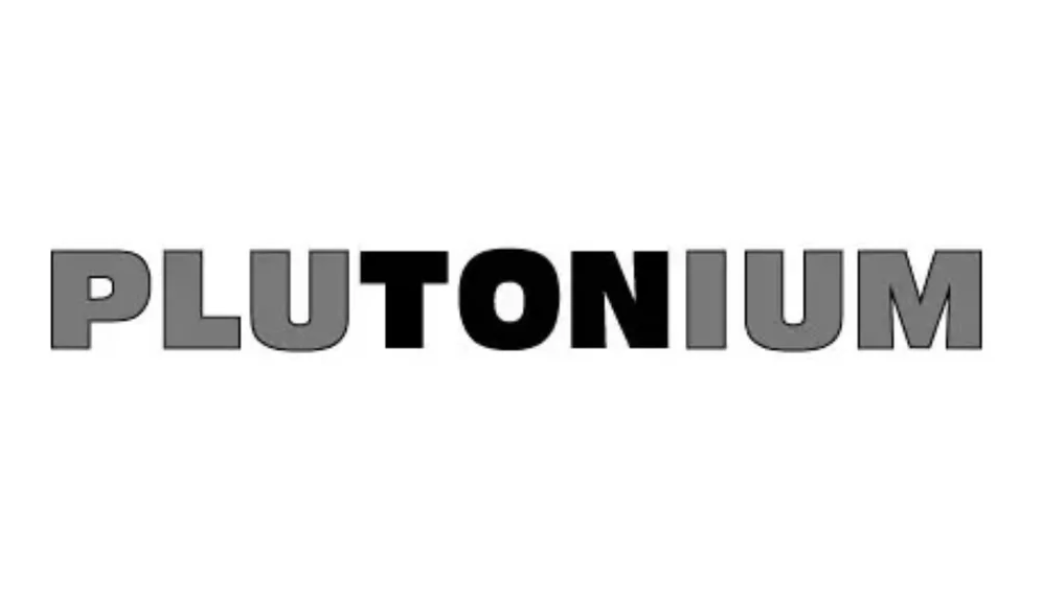}
    \caption{Heavy metal: Plutonium is a heavy metal and the word TON is emphasized in the text requiring logic for scientific classification and understanding of the image}
    \label{fig:plutonium}
\end{subfigure}
\hfill
\begin{subfigure}{0.45\linewidth}
    \includegraphics[width=\linewidth]{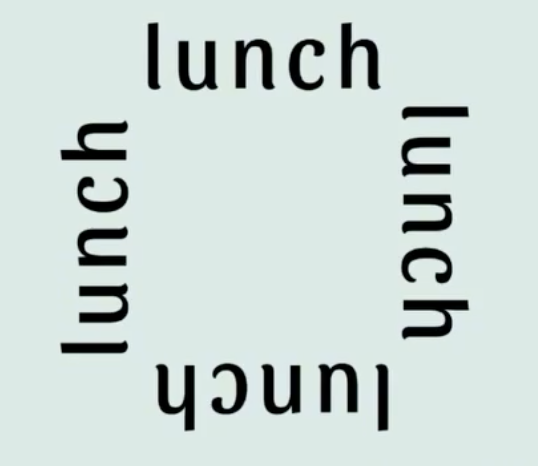}
    \caption{Boxed lunch/Lunch box: The words lunch are arranged in the shape of a square thus forming a boxed shape}
    \label{fig:lunch}
\end{subfigure}

\caption{Examples for errors in single-skill fixation.}
\label{fig:skill-fixation}
\end{figure}

Figure \ref{fig:pollinate} requires simultaneous spatial encoding (recognizing "PAUL" embedded in "EIGHT") and phonetic transformation (mapping "Paul-in-eight" to "pollinate"). GPT-o3 successfully integrates both skills across all strategies. Claude Sonnet, however, shows single-skill fixation: under ETS it recognizes "PAUL in EIGHT" spatially but answers "PAUL IS IN EIGHT" literally without attempting phonetic transformation. Under STE, Sonnet completely abandons spatial analysis and proposes "EGGPLANT" through invented anagram reasoning. The CG response proves most revealing: Sonnet recognizes the spatial embedding but attempts phonetic transformation toward "APPALLED," demonstrating that it can activate both skill types but cannot coordinate them toward a coherent solution. This suggests not missing capabilities but failure in cognitive orchestration.

Figure \ref{fig:eiffel-tower} presents a different multi-skill challenge: recognizing repeated "I FELL" text (visual composition) arranged in a tower-like descending pattern (spatial encoding) to represent the famous landmark. All models fail completely. GPT-o3 consistently interprets the visual as literal falling: "I fell down the stairs" across all strategies, correctly identifying the stair-like spatial arrangement but never considering phonetic transformation. Both Claude models show similar fixation—Sonnet proposes "WATERFALL" (recognizing cascading motion), while Opus suggests "HEAD OVER HEELS" (tumbling motion). The universal failure indicates that phonetically-driven solutions requiring non-literal sound mappings represent a particularly challenging reasoning mode. Models appear to exhaust literal interpretations of correctly identified visual patterns before considering whether textual elements might be phonetic proxies for entirely different words.

Figure \ref{fig:plutonium} requires quantitative logic (recognizing "TON" as a weight unit), and visual composition (identifying which letters are emphasized in "PLUTONIUM"). GPT-o3 successfully chains these inferences: plutonium is a metal, "TON" indicates weight/heaviness, therefore "heavy metal". Claude models consistently identify "TON" within "PLUTONIUM" but fail to make the connection. Opus's CG response exemplifies this: "The emphasized 'TON' in PLUTONIUM... suggests the answer relates to weight or heaviness" followed by "FINAL ANSWER: WEIGHT."

Figure \ref{fig:lunch} demonstrates successful multi-skill integration in contrast, though with interesting variations. The puzzle shows "lunch" repeated four times, each rotated 90° to form a square outline. This requires both spatial encoding (recognizing the box shape) and visual composition (understanding that multiple elements form a unified concept). GPT-o3 and Opus both solve this successfully, explicitly noting "the four 'lunch' words form a box shape." Sonnet, however, proposes "surrounded by lunch" under CG, correctly identifying the spatial relationship but choosing a descriptive phrase rather than the idiomatic "lunch box" or "boxed lunch." This reveals a subtler failure mode: executing all required reasoning but selecting non-conventional linguistic expressions.

These patterns suggest that models appear to select one dominant reasoning mode early in processing and struggle to simultaneously maintain alternative approaches, even when category labels explicitly signal that multiple skills are required.

\subsection{Evaluation metrics}
\label{app:eval-metrics}

\begin{figure}[H]
\centering
\includegraphics[width=0.8\linewidth]{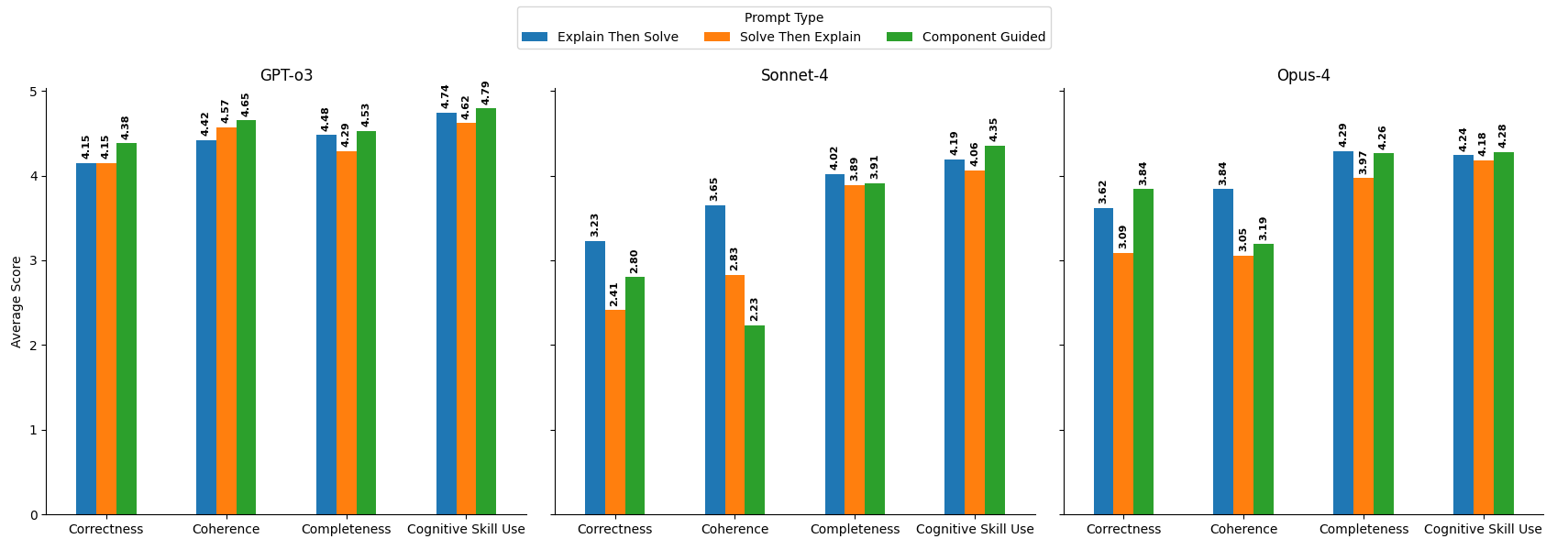}
\caption{Reasoning quality metrics by model and prompting strategy. CG prompting improves completeness and cognitive skill use across all models, even when correctness gains are modest.}
\label{fig:reasoning-scores}
\end{figure}

\end{document}